\documentclass[letterpaper, 10 pt, conference]{ieeeconf}  

\IEEEoverridecommandlockouts                              
\usepackage{siunitx}
\usepackage[table,x11names]{xcolor}
\usepackage{graphicx}
 \graphicspath{{./figures/}}

\usepackage{amsmath}
\usepackage{tcolorbox}
\usepackage{enumerate}
\let\labelindent\relax
\usepackage[shortlabels]{enumitem}\usepackage{amssymb}
\usepackage{latexsym}
\usepackage{url}
\usepackage{cite}
\usepackage{relsize}
\usepackage{multirow}
\usepackage{afterpage}
\usepackage{ifthen}
\usepackage{graphicx}
\usepackage{algpseudocode,algorithm,algorithmicx}
\usepackage{subfigure}
\usepackage{flushend}
\usepackage{epstopdf}
\usepackage{stackengine}
\usepackage{textcomp} 
\usepackage{pifont}
\newcommand{\cmark}{\ding{51}}%
\newcommand{\xmark}{\ding{55}}%
\usepackage{threeparttable,booktabs}
\usepackage{gensymb}
\usepackage{booktabs}
\usepackage{makecell}
\usepackage{hhline}
\usepackage{courier}
\usepackage{lipsum}
\usepackage{diagbox}
\usepackage{xcolor} 
\usepackage{xspace}
\providecommand{\FullStop}{\text{~\@.\xspace}}

\usepackage{xspace}

\makeatletter
\let\NAT@parse\undefined
\makeatother
\usepackage[bookmarks=false, linkcolor=blue, urlcolor=blue, citecolor=blue]{hyperref} 

\hypersetup{
    colorlinks=true,
    linkcolor=red,
    filecolor=magenta,      
    urlcolor=blue,
    pdfstartview={FitH},
    citecolor =blue
    }
\title{\LARGE \bf Grasp-Anything: Large-scale Grasp Dataset from Foundation Models}

\author{
A.D. Vuong$^{1}$, M.N. Vu$^{2, 3}$, H. Le$^{1}$, B. Huang$^4$, H.T.T. Binh$^5$, T. Vo$^6$, A. Kugi$^{2, 3}$, A. Nguyen$^7$
\thanks{$^1$ FPT Software AI Center, Vietnam {\tt anvd2@fpt.com}}
\thanks{$^2$ Automation \& Control Institute, TU Wien, Vienna, Austria 
}
\thanks{$^3$ AIT Austrian Institute of Technology, Vienna, Austria}
\thanks{$^4$ Imperial College London, UK
}
\thanks{$^5$ Hanoi University of Science and Technology, Vietnam}
\thanks{$^6$ Faculty of Mathematics and Statistics, Ton Duc Thang University, Ho Chi Minh City, Vietnam} 
\thanks{$^7$ Department of Computer Science, University of Liverpool, UK 
}}

\begin{document}

\newtheorem{problem}{Problem}
\newtheorem{lemma}{Lemma}
\newtheorem{theorem}[lemma]{Theorem}
\newtheorem{claim}{Claim}
\newtheorem{corollary}[lemma]{Corollary}
\newtheorem{definition}[lemma]{Definition}
\newtheorem{proposition}[lemma]{Proposition}
\newtheorem{remark}[lemma]{Remark}
\newenvironment{LabeledProof}[1]{\noindent{\it Proof of #1: }}{\qed}

\def\beq#1\eeq{\begin{equation}#1\end{equation}}
\def\bea#1\eea{\begin{align}#1\end{align}}
\def\beg#1\eeg{\begin{gather}#1\end{gather}}
\def\beqs#1\eeqs{\begin{equation*}#1\end{equation*}}
\def\beas#1\eeas{\begin{align*}#1\end{align*}}
\def\begs#1\eegs{\begin{gather*}#1\end{gather*}}

\newcommand{\poly}{\mathrm{poly}}
\newcommand{\eps}{\epsilon}
\newcommand{\e}{\epsilon}
\newcommand{\polylog}{\mathrm{polylog}}
\newcommand{\rob}[1]{\left( #1 \right)} 
\newcommand{\sqb}[1]{\left[ #1 \right]} 
\newcommand{\cub}[1]{\left\{ #1 \right\} } 
\newcommand{\rb}[1]{\left( #1 \right)} 
\newcommand{\abs}[1]{\left| #1 \right|} 
\newcommand{\zo}{\{0, 1\}}
\newcommand{\zonzo}{\zo^n \to \zo}
\newcommand{\zokzo}{\zo^k \to \zo}
\newcommand{\zot}{\{0,1,2\}}
\newcommand{\en}[1]{\marginpar{\textbf{#1}}}
\newcommand{\efn}[1]{\footnote{\textbf{#1}}}
\newcommand{\vecbm}[1]{\boldmath{#1}} 
\newcommand{\uvec}[1]{\hat{\vec{#1}}}
\newcommand{\thv}{\vecbm{\theta}}
\newcommand{\junk}[1]{}
\newcommand{\var}{\mathop{\mathrm{var}}}
\newcommand{\rank}{\mathop{\mathrm{rank}}}
\newcommand{\diag}{\mathop{\mathrm{diag}}}
\newcommand{\tr}{\mathop{\mathrm{tr}}}
\newcommand{\acos}{\mathop{\mathrm{acos}}}
\newcommand{\atantwo}{\mathop{\mathrm{atan2}}}
\newcommand{\SVD}{\mathop{\mathrm{SVD}}}
\newcommand{\quadf}{\mathop{\mathrm{q}}}
\newcommand{\linterp}{\mathop{\mathrm{l}}}
\newcommand{\sgn}{\mathop{\mathrm{sign}}}
\newcommand{\sym}{\mathop{\mathrm{sym}}}
\newcommand{\avg}{\mathop{\mathrm{avg}}}
\newcommand{\mean}{\mathop{\mathrm{mean}}}
\newcommand{\erf}{\mathop{\mathrm{erf}}}
\newcommand{\grad}{\nabla}
\newcommand{\R}{\mathbb{R}}
\newcommand{\defeq}{\triangleq}
\newcommand{\dims}[2]{[#1\!\times\!#2]}
\newcommand{\sdims}[2]{\mathsmaller{#1\!\times\!#2}}
\newcommand{\udims}[3]{#1}
\newcommand{\udimst}[4]{#1}
\newcommand{\com}[1]{\rhd\text{\emph{#1}}}
\newcommand{\ind}{\hspace{1em}}
\newcommand{\argmin}[1]{\underset{#1}{\operatorname{argmin}}}
\newcommand{\floor}[1]{\left\lfloor{#1}\right\rfloor}
\newcommand{\step}[1]{\vspace{0.5em}\noindent{#1}}
\newcommand{\quat}[1]{\ensuremath{\mathring{\mathbf{#1}}}}
\newcommand{\norm}[1]{\left\lVert#1\right\rVert}
\newcommand{\ignore}[1]{}
\newcommand{\specialcell}[2][c]{\begin{tabular}[#1]{@{}c@{}}#2\end{tabular}}
\newcommand*\Let[2]{\State #1 $\gets$ #2}
\newcommand{\algorithmicbreak}{\textbf{break}}
\newcommand{\Break}{\State \algorithmicbreak}
\newcommand{\ra}[1]{\renewcommand{\arraystretch}{#1}}

\renewcommand{\vec}[1]{\mathbf{#1}} 

\algdef{S}[FOR]{ForEach}[1]{\algorithmicforeach\ #1\ \algorithmicdo}
\algnewcommand\algorithmicforeach{\textbf{for each}}
\algrenewcommand\algorithmicrequire{\textbf{Require:}}
\algrenewcommand\algorithmicensure{\textbf{Ensure:}}
\algnewcommand\algorithmicinput{\textbf{Input:}}
\algnewcommand\INPUT{\item[\algorithmicinput]}
\algnewcommand\algorithmicoutput{\textbf{Output:}}
\algnewcommand\OUTPUT{\item[\algorithmicoutput]}

\maketitle
\thispagestyle{empty}
\pagestyle{empty}

\begin{abstract} 
Foundation models such as ChatGPT have made significant strides in robotic tasks due to their universal representation of real-world domains. In this paper, we leverage foundation models to tackle grasp detection, a persistent challenge in robotics with broad industrial applications. Despite numerous grasp datasets, their object diversity remains limited compared to real-world figures. Fortunately, foundation models possess an extensive repository of real-world knowledge, including objects we encounter in our daily lives. As a consequence, a promising solution to the limited representation in previous grasp datasets is to harness the universal knowledge embedded in these foundation models. We present Grasp-Anything, a new large-scale grasp dataset synthesized from foundation models to implement this solution. Grasp-Anything excels in diversity and magnitude, boasting 1M samples with text descriptions and more than 3M objects, surpassing prior datasets. Empirically, we show that Grasp-Anything successfully facilitates zero-shot grasp detection on vision-based tasks and real-world robotic experiments. Our dataset and code are available at \href{https://grasp-anything-2023.github.io}{https://grasp-anything-2023.github.io}.

\end{abstract}


\section{INTRODUCTION} \label{Sec: intro}
Grasp detection is a fundamental and long-standing research topic in robotics~\cite{pinto2016supersizing}. Establishing principles and techniques for grasp detection has enabled multiple applications such as manufacturing, logistics, and warehouse automation~\cite{sahbani2012overview}. Recent advances in deep learning have introduced effective avenues for the development of data-driven systems for robotic grasping~\cite{levine2018learning}. Numerous deep learning approaches have been proposed to address robotic grasping~\cite{kumra2020antipodal, song2020grasping, cao2023nbmod, Nguyen2023, bernardi2023learning}; however, they primarily focus on improving the neural network (i.e., \textit{model-centric} approach). While often excelling in the training progress, these model-centric strategies demonstrate unstable outcomes with different datasets, especially with real-world data~\cite{fang2023anygrasp}. Furthermore, Platt \textit{et al.}~\cite{platt2023grasp} contend that the outcomes conducted on physical robots strongly depend on the training data, underscoring the pivotal role of grasp data. Consequently, our paper explores the \textit{data-centric} approach, which aims to improve the quality of grasp data to achieve more robust generalization in grasp detection.



Over the years, many grasp datasets have been proposed (Table~\ref{table: grasp_dataset}). Nevertheless, current grasping datasets share common limitations. First, grasp datasets have been constrained by the limited number of objects~\cite{pourpanah2022review}. 
The restriction of the number of objects leads to dissimilarities among existing datasets and may introduce inconsistencies when transferring to real-life robotic applications~\cite{depierre2018jacquard, fang2023anygrasp}. Second, most of the current grasp datasets do not consider natural language descriptions for each scene arrangement, limiting human-robot interactions~\cite{wang2022ofa}. Finally, previous works make assumptions about scene arrangements resembling bin-like configurations~\cite{gilles2022metagraspnet} or lab-controlled environments~\cite{levine2018learning}, which diverges significantly from the complexity of natural settings~\cite{fang2023anygrasp}. To overcome the challenge, we aim to establish a new large-scale \textit{language-driven grasp dataset} that ideally covers unlimited scene arrangements in our daily lives.

\begin{figure}[t]
\centering
\includegraphics[width=0.99\linewidth]{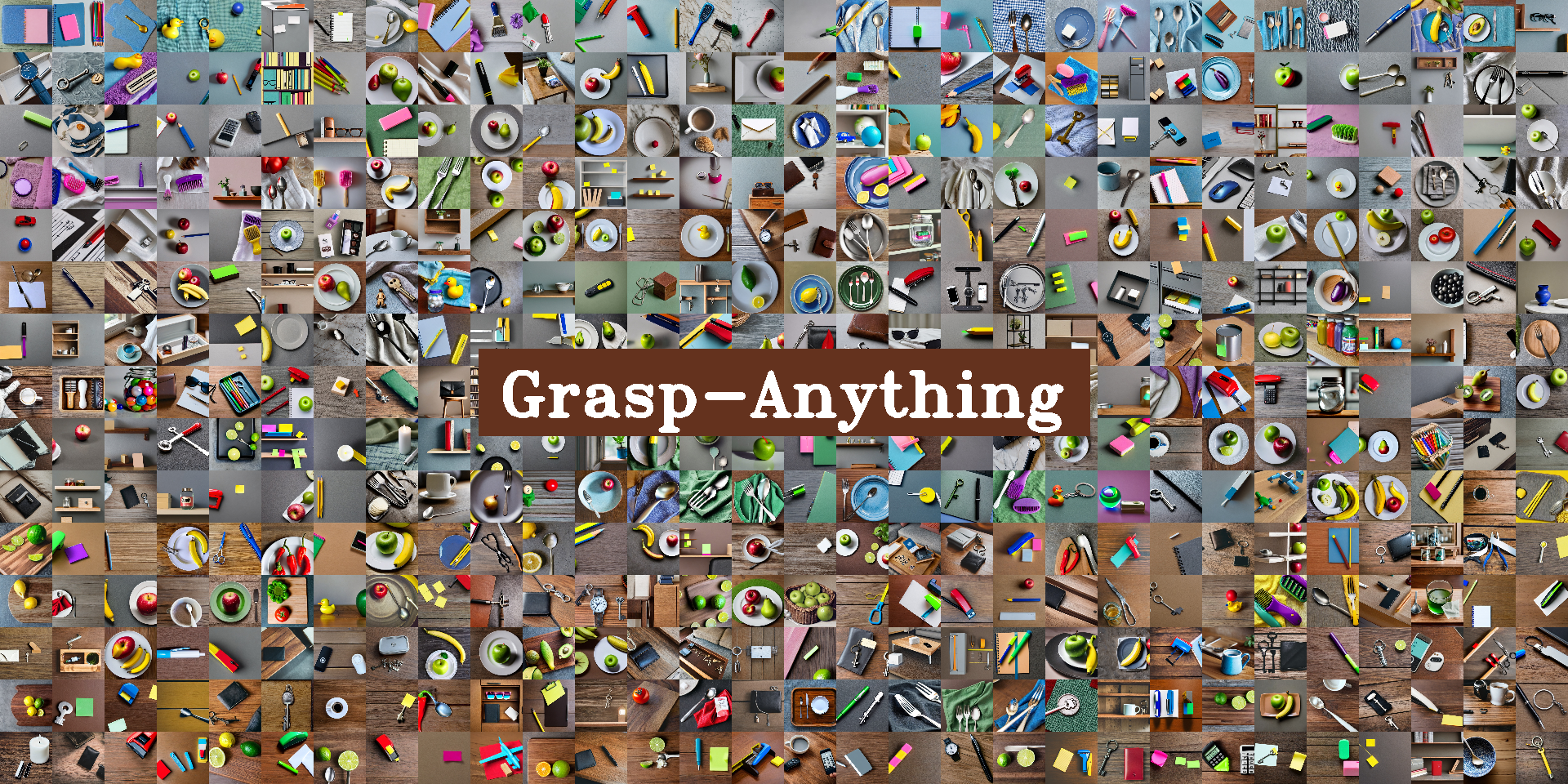}
\vspace{0.5ex}
\caption{We introduce Grasp-Anything, a new large-scale language-driven grasp dataset that universally covers objects in our daily lives by using knowledge from foundation models. 
}
\label{fig:intro}
\end{figure}

\begin{table*}[!t]
\centering
\caption{\label{table: grasp_dataset}Summary of the properties of publicly available grasp datasets.}
\vskip 0.1 in
\resizebox{\linewidth}{!}{
\begin{tabular}{|l|c|c|c|c|c|c|c|c|c|c|}
\hline
\rowcolor{lightgray} Dataset                                                           & \begin{tabular}[c]{@{}c@{}}Grasp \\ Label\end{tabular} & \begin{tabular}[c]{@{}c@{}}Data \\ Representation\end{tabular} & \begin{tabular}[c]{@{}c@{}}Multi-\\ object?\end{tabular} & \begin{tabular}[c]{@{}c@{}}Grasps\\ per object\end{tabular} & \textbf{\begin{tabular}[c]{@{}c@{}}Num. \\ objects\end{tabular}} & \begin{tabular}[c]{@{}c@{}}Num.\\ grasps\end{tabular} & \textbf{\begin{tabular}[c]{@{}c@{}}Num. \\ samples\end{tabular}} & \begin{tabular}[c]{@{}c@{}}Data\\ type\end{tabular} & Label annotation & \multicolumn{1}{l|}{\begin{tabular}[c]{@{}c@{}}Scene\\ description?\end{tabular}} \\ \hline
Cornell~\cite{jiang2011efficient}           & Rec.                                                   & Image                                                          & \xmark                                    & 33                                                         & 240                                                                  & 8019                                                        & 1035 & Real                                                & Human                                                      & \xmark                                                                \\ \hline
Pinto \textit{et al}.~\cite{pinto2016supersizing}    & Rec.                                                   & Image                                                          & \cmark                                    & 1                                                          & 150                                                                  & 50K                                                         & 50K & Real                                                & Trial-and-error                                            & \xmark                                                                \\ \hline
Levine \textit{et al}.~\cite{levine2018learning}     & Rec.                                                   & Image                                                          & \cmark                                    & -                                                          & -                                                                    & 800K                                                        & - & Real                                                & Trial-and-error                                            & \xmark                                                                \\ \hline
Dex-net~\cite{mahler2017dex}                & Rec.                                                   & Image                                                              & \xmark                                    & 100                                                        & 1500                                                                 & 6.7M                                                        & - & Sim                                                 & Analysis                                                          & \xmark                                                                \\ \hline
Jacquard~\cite{depierre2018jacquard}        & Rec.                                                   & Image                                                          & \xmark                                    & 100                                                        & 11K                                                                  & 1.1M                                                        & 54K & Sim                                                 & Sim. (PyBullet~\cite{coumans2019})                                                 & \xmark                                                                \\ \hline
VMRD~\cite{zhang2019roi}                    & Rec.                                                   & Image                                                            & \cmark                                    & 20                                                         & 15K                                                                   & 100K                                                        & 4683  & Real                                                & Human                                                      & \xmark                                                                \\ \hline
OCID-grasp~\cite{ainetter2021end}                    & Rec.                                                   & Image                                                            & \cmark                                    & 6-7                                                         & $\sim$40                                                                  & 75K                                                       & 11K & Real                                                & Human                                                      & \xmark                                                                \\ \hline

VR-Grasping-101~\cite{yan2018learning}      & 6-DoF                                                  & Image                                                          & \xmark                                    & 10-20                                                      & 101                                                                  & 151K                                                        & -  & Sim                                                 & Trial-and-error                                            & \xmark                                                                \\ \hline
YCB-Video~\cite{xiang2017posecnn}           & None                                                   & Image                                                          & \cmark                                    & None                                                       & 21                                                                   & None                                                        & 134K  & Real                                                & None                                                       & \xmark                                                                \\ \hline
GraspNet~\cite{fang2020graspnet}            & 6-DoF                                                  & Image                                                          & \cmark                                    & 3-9M                                                       & 88                                                                   & 1.2B                                                        & 97K & Real                                                & Analysis                                                          & \xmark                                                                \\ \hline
Kappler \textit{et al}~\cite{kappler2015leveraging} & 6-DoF                                                  & Point Cloud                                                 & \xmark                                    & 500                                                        & 80                                                                   & 300K                                                        & 700 & Sim                                                 & Analysis                                                          & \xmark                                                                \\ \hline
6-DOF GraspNet~\cite{mousavian20196}        & 6-DoF                                                  & Point Cloud                                                 & \xmark                                    & 34K                                                        & 206                                                                    & 7M                                                          & 206 & Sim                                                 & Sim. (FleX~\cite{macklin2014unified})                                                 & \xmark                                                                \\ \hline
Epper \textit{et al}.~\cite{eppner2019billion}       & 6-DoF                                                  & Point Cloud                                                 & \xmark                                    & 47.8M                                                      & 21                                                                   & 1B                                                          & 21 & Sim                                                 & Sim. (FleX~\cite{macklin2014unified})                                                 & \xmark                                                                \\ \hline
EGAD!~\cite{morrison2020egad}               & 6-DoF                                                  & Point Cloud                                                 & \xmark                                    & 100                                                        & 2231                                                                 & 233K                                                        & 2231  & Sim                                                 & Analysis                                                          & \xmark                                                                \\ \hline
ACRONYM~\cite{eppner2021acronym}            & 6-DoF                                                  & Point Cloud                                                 & \cmark                                    & 2000                                                       & 8872                                                                 & 17.7M                                                       & -  & Sim                                                 & Sim. (FleX~\cite{macklin2014unified})                                                 & \xmark                                                                \\ \hline
MetaGraspNet~\cite{gilles2022metagraspnet}            & 6-DoF                                                  & Point Cloud                                                 & \cmark                                    &    5K                                                    & 82                                                                 & -                                                       & 217K & Sim                                                 & Sim. (Isaac Sim~\cite{nvidia_omniverse})                                                 & \xmark                                                                \\ 
\hhline{-----------}\hhline{-----------}
Grasp-Anything (Ours)                                             & Rec.                                                    & Image                                                            & \cmark                                    & $\sim$200                                                  & $\sim$3M                                                                 & $\sim$600M                                                       & 1M         & Synth.                                           & Analysis                                                          & \cmark                                                                \\ \hline
\end{tabular}}
\end{table*}

Recently, we have witnessed the applications of utilizing foundation models, such as large language model (LLM)~\cite{openai2021chatgpt} or text-to-image (T2I) model~\cite{ rombach2022high}, across various domains of robotics research~\cite{bucker2022reshaping}. Foundation models have demonstrated remarkable encouragements in various tasks~\cite{vemprala2023chatgpt}, for example, task and motion planning~\cite{singh2022progprompt}, manipulation~\cite{kapelyukh2023dall}, visual-and-language navigation~\cite{wang2023voyager,dinh2023habicrowd}, and scene understanding~\cite{peng2023openscene}. The utilization of large-scale foundation models has facilitated the integration of omniscient knowledge into robotic systems~\cite{liu2023llm+}, overcoming the challenges faced by traditional methods in robustly modeling unstructured and novel environments~\cite{ren2023robots}. Inspired by these phenomena, we hypothesize that it is possible to apply panoptic knowledge from large foundation models to ideally generate an unlimited number of objects and, therefore, serve as a foundation to synthesize a grasp dataset that universally covers possible objects and arrangements that come into existence.

We introduce \textbf{Grasp-Anything}, a new large-scale dataset for grasp detection. Unlike existing grasp datasets~\cite{jiang2011efficient, depierre2018jacquard, pinto2016supersizing, levine2018learning, kappler2015leveraging, xiang2017posecnn, nguyen2017object,mahler2017dex, mousavian20196, eppner2019billion, fang2020graspnet, morrison2020egad, eppner2021acronym, cao2023nbmod, song2020grasping} that are limited to a predefined set of objects and arrangements, our dataset offers an extensive range of objects and closely replicates real-world scenarios in natural environments, thus alleviating the generalization issues~\cite{driess2023palm}. We empirically demonstrate that Grasp-Anything facilitates zero-shot learning in both computer vision and robotic aspects of grasp detection. In the vision aspect, our findings confirm that the performances of grasp detection baselines trained by Grasp-Anything improve significantly over related datasets. In the robotic aspect, we demonstrate that our large-scale dataset can be directly applied to real robot systems to improve the grasping task. In summary, our contributions are as follows:

\begin{itemize}
    \item We leverage knowledge from foundation models to introduce Grasp-Anything, a new large-scale dataset with 1M (one million) samples and 3M objects, substantially surpassing prior datasets in diversity and magnitude.
    \item We benchmark zero-shot grasp detection on various settings, including real-world robot experiments. The results indicate that Grasp-Anything effectively supports zero-shot grasp detection in light of its comprehensive representation of real-world scene arrangements.
\end{itemize}
\section{Related Work} \label{Sec: related_work}
\textbf{Grasp Datasets.} Several grasp datasets have been introduced recently~\cite{newbury2023deep} (see Table~\ref{table: grasp_dataset}). Many factors can be considered when designing a grasp dataset, such as data representation (RGB-D or 3D point clouds), grasp labels (rectangle-based or 6-DoF), and quantity~\cite{pourpanah2022review}. Notably, a key distinction between our Grasp-Anything dataset and its counterparts lies in its \textit{universality}. 
While a limited selection of objects constrains existing benchmarks, our dataset is designed to encompass a diverse spectrum of objects observed in our natural lives. In addition, our dataset incorporates natural settings for object arrangements, thereby distinguishing itself from prior works where object configurations are more strictly controlled~\cite{platt2023grasp}. Grasp-Anything outperforms other benchmarks in both the number of objects and the number of samples. 

\textbf{Grasp Detection.} 
Deep learning techniques have demonstrated notable achievements in grasp detection. Lenz \textit{et al.}~\cite{lenz2015deep} present one of the first works utilizing deep learning to detect grasp pose. Subsequently, learning-based approaches~\cite{ yan2018learning, liang2019pointnetgpd, jiang2021synergies, wen2022catgrasp, ainetter2021end, kumra2020antipodal, cao2023nbmod} have gained prominence as the most widely utilized solution for grasp detection. Despite the extensive research on deep learning methods for robotic grasp, it is still challenging to apply to real-world grasping applications~\cite{platt2023grasp}. This is primarily attributed to the limited size and diversity of existing datasets for robotic grasp~\cite{gilles2022metagraspnet}. 
On the other hand, zero-shot grasp detection is a promising approach for utilizing large-scale datasets. Liu \textit{et al.}~\cite{liu2023digital} consider the zero-shot grasp detection problem. However, they mainly focus on sim2real applications rather than benchmarking the zero-shot feasibility of existing datasets. Therefore, we are motivated to develop a large-scale benchmark for grasp detection, enabling the successful grasp of universal objects we encounter in real-life scenarios.


\textbf{Foundation Models for Robotic Applications.} Different attempts to incorporate foundation models into robotic applications have been proposed~\cite{wang2023voyager, yang2023pave}. For instance, Kapelyukh \textit{et al.}~\cite{kapelyukh2023dall} introduce a framework based on DALL-E~\cite{ramesh2022hierarchical} to solve practical rearrangement tasks. Pretrained image in-painting models have also been utilized in~\cite{yu2023scaling} for data augmentation. Although applying foundation models to robotic systems has become an inevitable trend~\cite{vemprala2023chatgpt}, it still bears several uncertainties~\cite{liu2023llm+}. First, these models need to be improved in functional competence~\cite{liu2023llm+}, impeding their ability to solve novel planning problems that require an understanding of how the world operates~\cite{mahowald2023dissociating}. In addition, prior approaches often exhibit drawbacks in terms of scope, limited functionalities, or an open-loop nature that does not allow for fluid interactions and corrections based on user feedback~\cite{vemprala2023chatgpt}. Recognizing these limitations, we design a data-centric approach based on LLM and T2I models, focusing on generating large-scale data for robotic grasping within this paper.

\section{The Grasp-Anything Dataset} \label{Sec: method}

\begin{figure*}[!ht]
    \centering
    \includegraphics[width=1.02\linewidth]{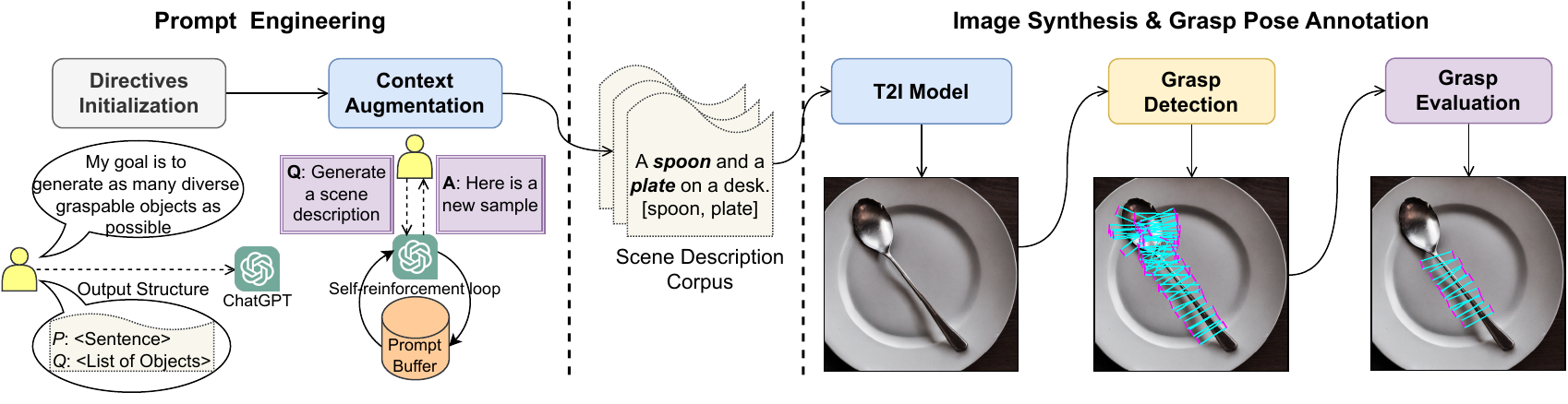}
    \caption{\textbf{Dataset creation pipeline.} 
    }
    \label{fig:data-generation}
\end{figure*}

Fig.~\ref{fig:data-generation} shows the overview of the procedure to generate our Grasp-Anything dataset. We first perform prompt engineering to generate scene descriptions and utilize foundation models to generate images from these text prompts. The grasp poses are then automatically generated and evaluated. We represent grasp poses as 2D
rectangles~\cite{jiang2011efficient} as in many previous works due to the simplicity and compatibility with real-world parallel plate grippers~\cite{depierre2018jacquard}.

\subsection{Scene Generation}
\textbf{Prompt Engineering.} To tackle the challenge of generating a universal set of objects, we utilize ChatGPT~\cite{openai2021chatgpt} and perform a prompt engineering technique~\cite{reynolds2021prompt} to guide ChatGPT to generate diverse scene descriptions. 

\textit{i) Directives Initialization.} The concept of `directives' refers to the configuration of the goal for ChatGPT to match the user preference~\cite{wang2023voyager}. In our context, the goal for ChatGPT is to generate a diverse set of scene descriptions that cover a large proportion of objects. Therefore, we initiate directives for ChatGPT using the following conversation

\vspace{-1.2ex}
\begin{tcolorbox}
\vspace{-1.2ex}
\begin{itemize}[leftmargin=*]
    \item[\textbf{Q:}] ``Imagine you are helping me to generate a corpus of scene descriptions, each condensed to a single sentence. My goal is to generate as many diverse graspable objects as possible. Each sentence must be distinct and should contain at least two objects."
    \item[\textbf{A:}] ``Sure, I'd be happy to help you generate a large corpus of scene descriptions with diverse objects."
\end{itemize}
\vspace{-1.4ex}
\end{tcolorbox}
\vspace{-1.0ex}

We then set up the output template for ChatGPT as follows

\vspace{-1.2ex}
\begin{tcolorbox}
\vspace{-1.2ex}
\begin{itemize}[leftmargin=*]
    \item[\textbf{Q:}] ``The template for each sentence contains two parts. The first part is the sentence with the structure as \texttt{\textless Obj\_1\textgreater\textless Obj\_2\textgreater$\ldots$\textless Verb\textgreater \textless Container\textgreater}. The second part is the list of extracted objects from the sentence \texttt{[\textless Obj\_1\textgreater\textless Obj\_2\textgreater$\ldots$]}."
    \item[\textbf{A:}] ``Understood! Let's generate the first sentence$\ldots$"
\end{itemize}
\vspace{-1.4ex}
\end{tcolorbox}
\vspace{-1.0ex}
This template guarantees that each generated prompt has two components: the text describing the scene arrangement and a list indicating graspable objects in the text.

\textit{ii) Context Augmentation.} Directives guide ChatGPT at a high level for scene descriptions but do not guarantee long-term quality due to hallucination~\cite{vemprala2023chatgpt}. To ensure consistent quality over a long time horizon, we employ a context augmentation by creating a self-reinforcing loop for ChatGPT. We initialize a prompt buffer to store generated prompts, with the first 50 samples \textit{manually} assigned. We sample 10-15 scene descriptions from the prompt buffer each time and input them to ChatGPT. An example of this process is given in the following example

\vspace{-1.2ex}
\begin{tcolorbox}
\vspace{-1.2ex}
\begin{itemize}[leftmargin=*]
    \item[\textbf{Q:}] ``Based on sample scene descriptions, generate a new scene description with a similar structure."
    \item[\textbf{A:}] ``Certainly! Here's a new scene description: A spoon and a plate on a desk. [spoon, plate]"
\end{itemize}
\vspace{-1.4ex}
\end{tcolorbox}
\vspace{-1.0ex}

The new sample is then appended to the prompt buffer. We repeat the process until 1M scene descriptions are generated.



\textbf{Image Synthesis.} Given the scene descriptions generated by ChatGPT, we use Stable Diffusion 2.1~\cite{rombach2022high} to generate images that align with the scene descriptions. We then gather instance segmentation masks for every object that appeared in the grasp list using the state-of-the-art visual grounding and instance segmentation models (OFA~\cite{wang2022ofa} and Segment-Anything~\cite{kirillov2023segment}). At the end of the image synthesis stage, we obtain a grounding mask for each referenced object. 


\begin{figure}[!ht]
\centering
\subfigure[Segmentation mask]{\label{fig:convex_hull_construction}\includegraphics[height=35mm]{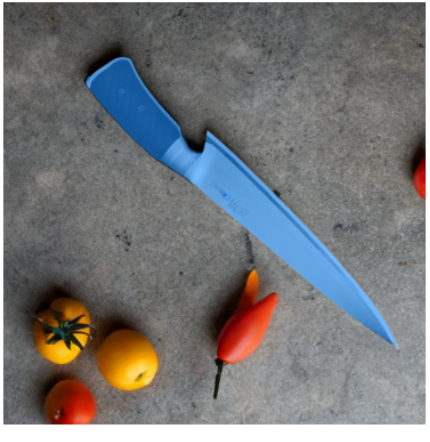}}
\hspace{0ex}
\subfigure[Torque of a grasp pose]{\label{fig:grasp_pose_evaluation}\includegraphics[height=35mm]{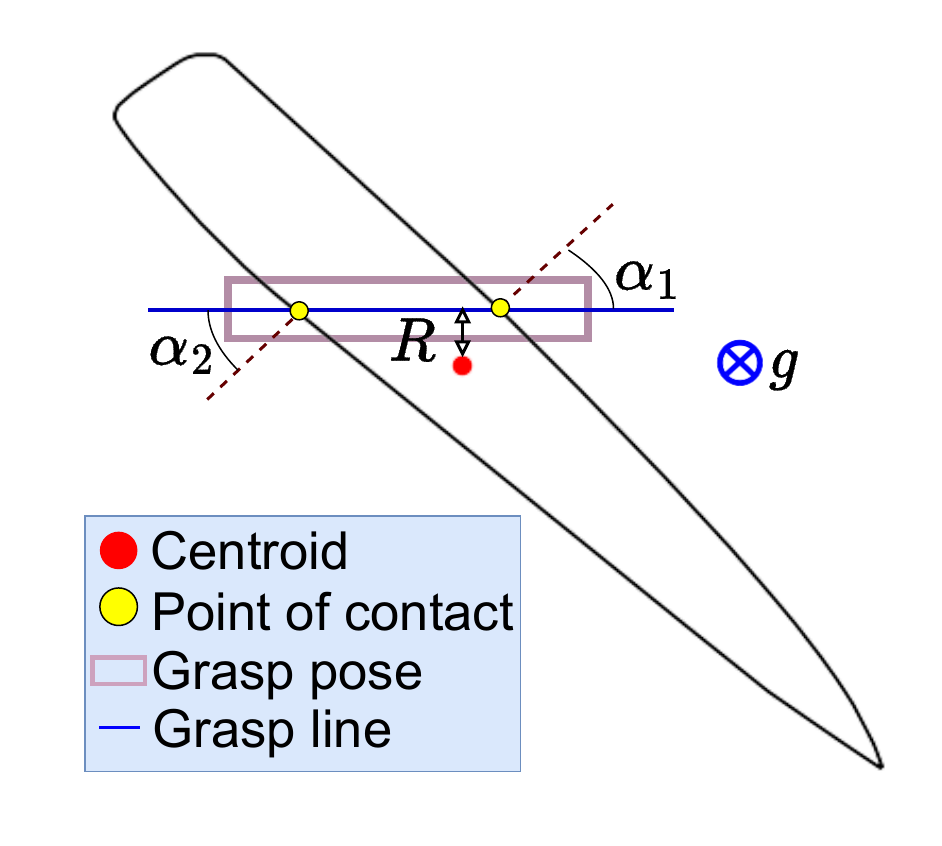}}
\vspace{1ex}
\caption{\textbf{Grasp pose evaluation}. We construct a convex hull for each object segmentation mask and use the theory of~\cite{kamon1996learning} to determine feasible grasps.}
\label{fig:intro}
\end{figure}

\begin{figure*}[t]
\centering
\subfigure[Grasp-Anything sorted by LVIS categories.]{\label{fig:lvis-label}\includegraphics[width=0.32\linewidth]{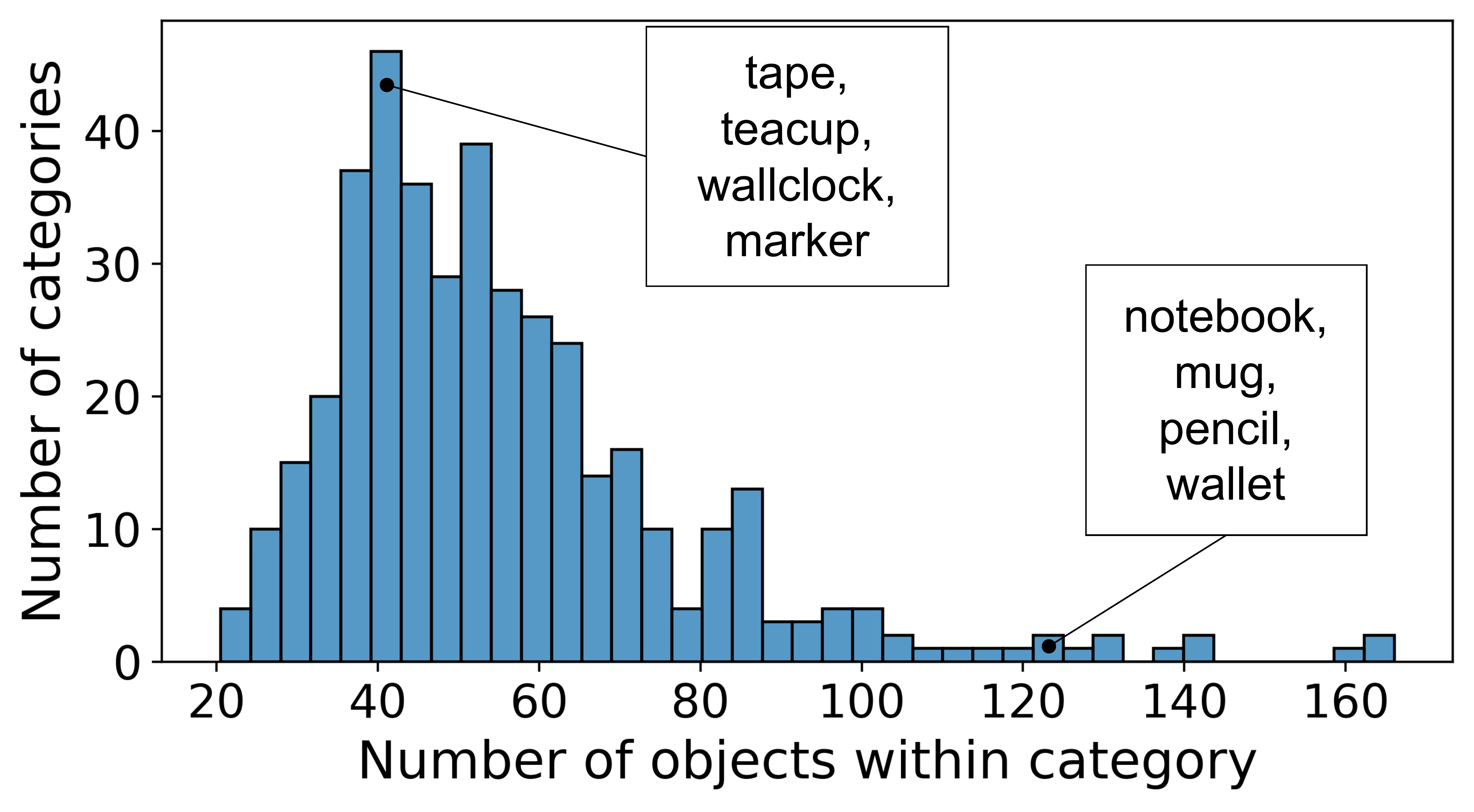}}
\hspace{0.2ex}
\subfigure[Number of objects (in log-scale) comparison.]{\label{fig:num_objs_comparison}\includegraphics[width=0.32\linewidth]{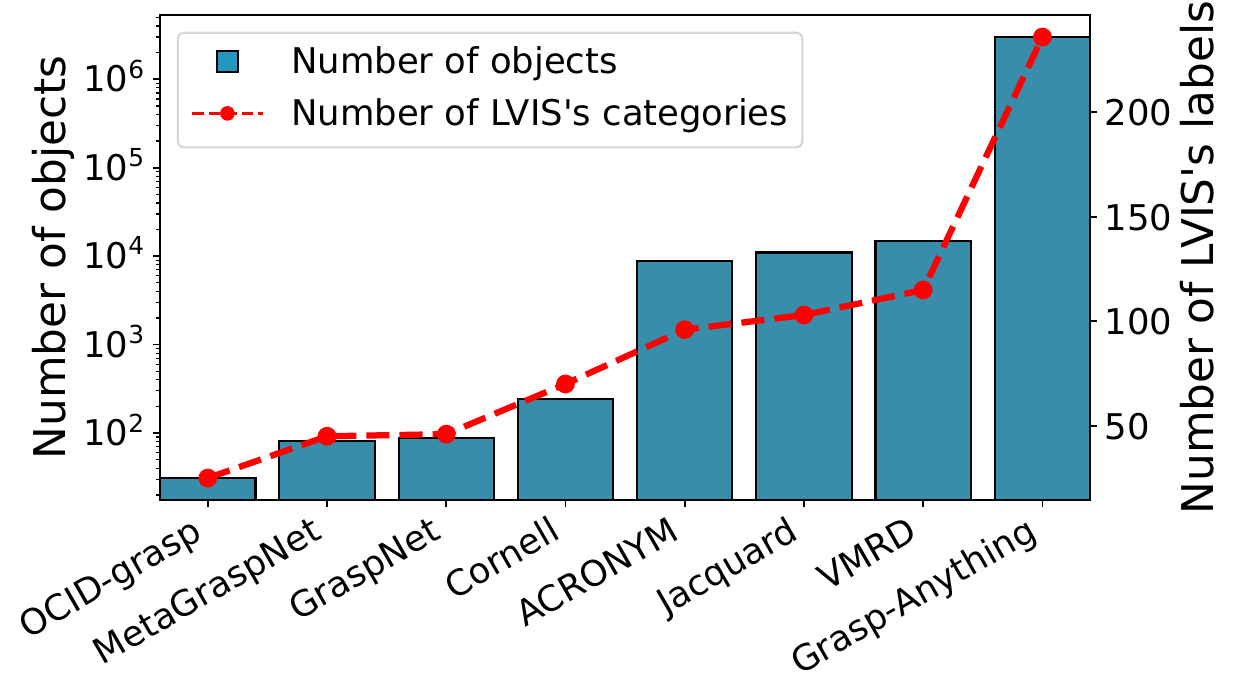}}
\hspace{0.2ex}
\subfigure[Grasp-Anything's POS tags.]{\label{fig:pos-tag-distribution}\includegraphics[width=0.32\linewidth]{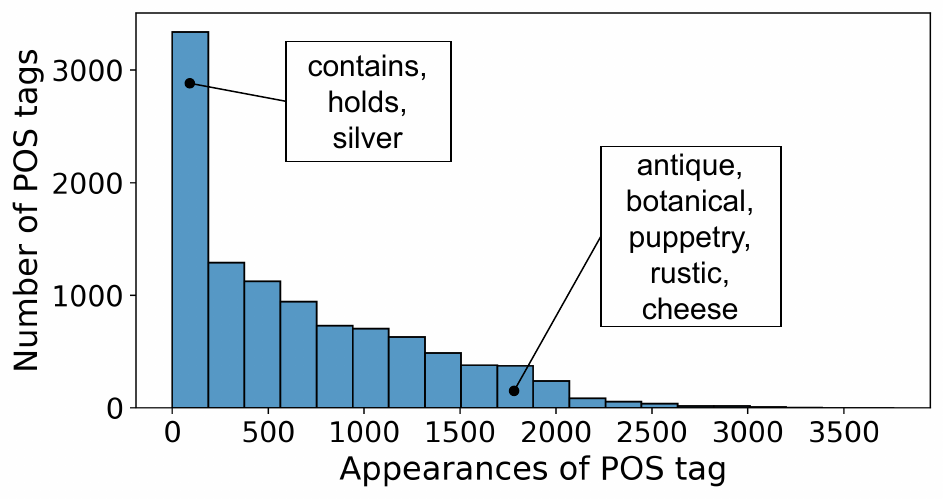}}
\vspace{1ex}
\caption{\textbf{Dataset Statistics}. We provide statistics on (a) the number of categories, (b) the number of POS tags, and (c) a comparison of number of objects.}
\vspace{-1.35ex}
\end{figure*}

\subsection{Grasp Pose Annotation}

\begin{figure}[!ht]
    \centering
    \includegraphics[width=1.02\linewidth]{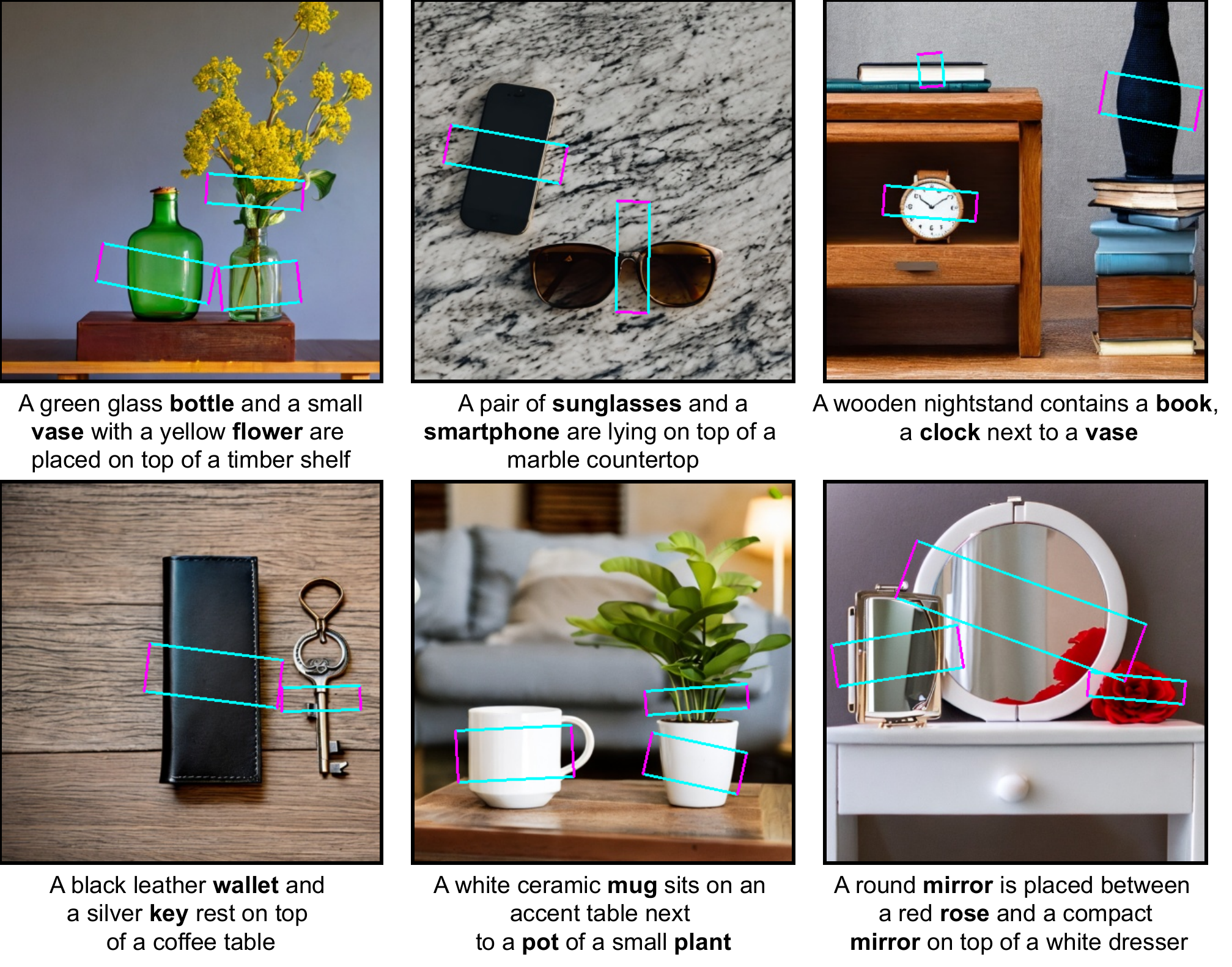}
    \caption{\textbf{Samples from Grasp-Anything.} For simplicity, we only display the grasp with the highest $\tilde{\mathcal{T}}$ for each graspable object (marked in bold). 
    }
    \label{fig:grasp-anything-sample}
\end{figure}

To annotate grasp poses for each object in the grasp list of the scene description, we employ a pretrained model RAGT-3/3~\cite{cao2023nbmod}. Since the candidate poses may not always be accurate, we use a traditional method by Kamon \textit{et al.}~\cite{kamon1996learning} to further evaluate grasp poses. 
More specifically, we determine the grasp quality of each pose by calculating the net torque, denoted as $\mathcal{T}$, associated with the grasp as follows

\begin{equation}\label{eq: net_torque}
    \mathcal{T} = \underbrace{\left(\tau_1+\tau_2\right)}_\textrm{Resistance} - \underbrace{RMg}_\textrm{Torque}\FullStop
\end{equation}
The resistance at each contact point, denoted as $\tau_i$, can be computed by $\tau_i = K\mu_sF\cos\alpha_i, \forall i\in\{1, 2\}$. In Equation~\eqref{eq: net_torque}, the terms $M, g, K, \mu_s, F$ correspond to the object's mass, gravitational acceleration, geometrical characteristics of the contact area, coefficient of static friction, and the applied force, respectively. These parameters are assumed to be constant across all grasps. Consequently, each grasp pose is characterized by three variables: $R, \alpha_1, \alpha_2$, as depicted in Fig.~\ref{fig:grasp_pose_evaluation}. Due to the impracticality of explicitly determining the physical terms for each object~\cite{caldera2018review}, the computation of $\mathcal{T}$ becomes infeasible without knowing the physical parameters $M, K, \mu_s$. We employ the following concept to address the challenges posed by physical constraints

\begin{equation}\label{eq: modified_net_torque}
    \tilde{\mathcal{T}} = \dfrac{\cos\alpha_1 + \cos\alpha_2}{R}\FullStop
\end{equation}
The term $\tilde{\mathcal{T}}$ indicates the ratio between the resistance and the torque. By examining Equations~\eqref{eq: net_torque} and~\eqref{eq: modified_net_torque}, we verify that the original net torque $\mathcal{T}$ is correlated to $\tilde{\mathcal{T}}$. In cases a grasp results in fewer than two contact points, we assign each missing term $\cos\alpha_i$ a value of $-1$. Antipodal grasps tend to yield greater positive values of $\tilde{\mathcal{T}}$ compared to non-antipodal grasps~\cite{chen1993finding}. Thus, grasp qualities can be ranked based on $\tilde{\mathcal{T}}$. We define grasps with positive $\tilde{\mathcal{T}}$ values as positive grasps, while the remaining grasps are considered negative. We remark that our grasp evaluation procedure can only be applied when the segmentation mask for each object is available to compute its convex hull. Some examples of our dataset can be found in Fig.~\ref{fig:grasp-anything-sample}.

\subsection{Grasp-Anything Statistics}\label{subsec: data-stat}


\begin{table*}[!ht]
\centering
\renewcommand
\tabcolsep{4.5pt}
\hspace{1ex}
\caption{\label{table: base-to-new-grasp-detection} Base-to-new grasp detection result.}
\vskip 0.1 in
\begin{tabular}{@{}rccccccccccccccccc@{}}
\toprule
  & \multicolumn{3}{c}{\textbf{Grasp-Anything (ours)}} & \multicolumn{3}{c}{\textbf{Jacquard~\cite{depierre2018jacquard}}} & \multicolumn{3}{c}{\textbf{Cornell~\cite{jiang2011efficient}}} & \multicolumn{3}{c}{\textbf{VMRD~\cite{zhang2019roi}}} & \multicolumn{3}{c}{\textbf{OCID-grasp~\cite{ainetter2021end}}}\\
\cmidrule(lr){2-4}\cmidrule(lr){5-7}\cmidrule(lr){8-10}\cmidrule(lr){11-13}\cmidrule(lr){14-16}
Baseline &  
Base &  New & H & Base &  New & H & Base &  New & H & Base &  New & H & Base &  New & H \cr 
\midrule
GR-ConvNet~\cite{kumra2020antipodal} & \textbf{0.74} &  \textbf{0.61} & \textbf{0.67} & \textbf{0.88} & \textbf{0.66} & \textbf{0.75} & 0.98 & 0.74 & 0.84 & \textbf{0.77} & \textbf{0.64} & \textbf{0.70} & \textbf{0.86} & \textbf{0.67} & \textbf{0.75} \\

Det-Seg-Refine~\cite{ainetter2021end} & 0.64 & 0.59 & 0.61 & 0.86 &  0.60 & 0.71 & \textbf{0.99} &  \textbf{0.76} & \textbf{0.86} & 0.75 & 0.60 & 0.66 & 0.80 & 0.62 & 0.70\\

GG-CNN~\cite{morrison2018closing} & 0.71 & 0.59 & 0.64 & 0.78 & 0.56 & 0.65 & 0.96 & 0.75 & 0.84 & 0.69 & 0.53 & 0.59 & 0.71 & 0.63 & 0.67 \\
\bottomrule
\end{tabular}
\end{table*}

\textbf{Number of Categories.} To assess the diversity of object categories in our dataset, we conduct the same methodology as in~\cite{deitke2023objaverse}. We leverage 300 categories of LVIS~\cite{gupta2019lvis} dataset and identify 300 candidate objects from Grasp-Anything for each category using a pretrained model~\cite{zhong2022regionclip}. Subsequently, we select a subset of 90,000 objects and filter out objects not semantically aligned with their assigned categories. Each category is considered significant if it has more than 40 objects. The outcomes of this analysis are showed in Fig.~\ref{fig:lvis-label}. With this setup, we observe that our Grasp-Anything spreads over 236 categories of the LVIS dataset. We apply the same procedure to other datasets and show the comparison in Fig.~\ref{fig:num_objs_comparison}. It is noteworthy that VMRD~\cite{zhang2019roi} dataset is the runner-up and only has 115 LVIS categories, compared to 236 categories in our dataset.

\textbf{Number of Objects.} We visualize the number of objects (in log-scale) of Grasp-Anything and other grasp datasets in Fig.~\ref{fig:num_objs_comparison}. From this figure, we can see that our Grasp-Anything dataset has a significantly larger number of objects than other datasets.


\textbf{Number of POS tags.} To categorize words in a text according to their grammatical roles and syntactic functions, we extract the POS tags~\cite{tehseen2023neural} in our dataset and visualize them in Fig.~\ref{fig:pos-tag-distribution}. Our scene descriptions corpus utilizes a wide range of words to describe scene arrangements. 
We provide about 1.5M POS tags, with 35\% being nouns, 20\% being adjectives, 7\% being verbs, and the remainder being other POS tags such as prepositions.

\begin{figure}[!ht]
    \centering
    \includegraphics[width=\linewidth]{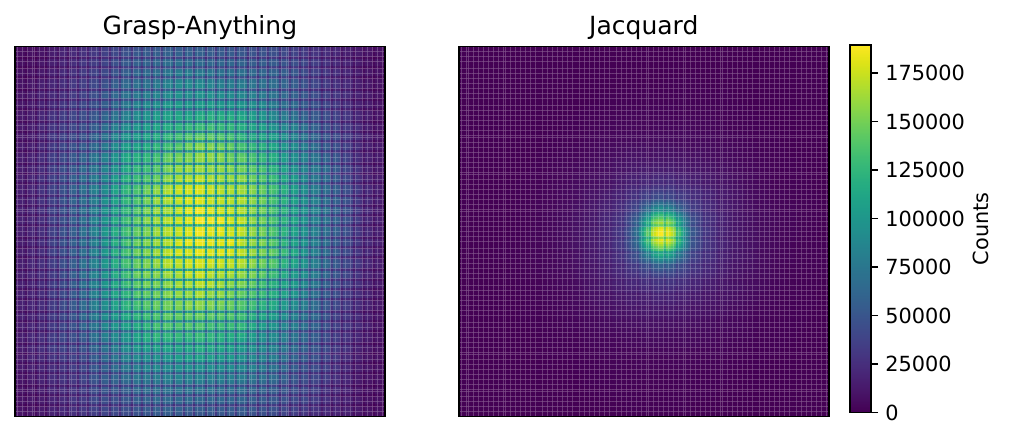}
    \caption{\textbf{Object shape heatmap visualization.}}
    \label{fig: shape-visualization}
\end{figure}

\textbf{Object Shape Distribution.}
Fig.~\ref{fig: shape-visualization} compares object shape distributions between Grasp-Anything and Jacquard. We randomly select $5000$ objects from each dataset and extract the $(x, y)$ coordinates associated with pixels lying in the interior of each object. We then aggregate all coordinates and combine them to create heatmaps. These heatmaps are normalized by each dataset's image resolution. We can see that objects in Grasp-Anything span over a greater area than Jacquard's, indicating a greater degree of shape diversity.



\subsection{How will Grasp-Anything be helpful to the community?}
Given the large-scale nature and its multi-modalities, such as text prompts, images, and segmented masks, we hope that our new dataset can drive future research in these topics:
\begin{itemize}[leftmargin=*]
    \item \textit{Grasp Detection}: Although many grasp datasets have been established~\cite{jiang2011efficient, depierre2018jacquard, pinto2016supersizing, levine2018learning, platt2023grasp, gilles2022metagraspnet}, we demonstrate that Grasp-Anything accommodates a broader range of objects and more natural scene settings; therefore, our dataset can advance more grasp detection research, especially on zero-shot grasp detection and domain adaptation.
    
    \item \textit{Language-driven Grasping}: Language-driven grasping detection is a promising research area with industrial applications~\cite{ahn2022can, xu2023joint, yang2023pave}. Grasp-Anything supports diverse scene descriptions, enabling large-scale training to align natural language with grasp detection.
\end{itemize}

Furthermore, we believe Grasp-Anything can be useful in related tasks such as sim2real grasping~\cite{horvath2022sim2real}, human-robot interaction~\cite{mivseikis2020lio}, or language-driven mobile manipulation~\cite{driess2023palm}.

 \graphicspath{./figures/}

\section{Experiments} \label{Sec: experiments}
We conduct experiments to validate the usefulness of our Grasp-Anything dataset and answer \textit{two questions}: \textit{i)} Can Grasp-Anything serve as a challenging dataset for grasp detection? and \textit{ii)} Since our dataset is generated by the foundation models, will it work on real robot experiments?




\subsection{Zero-shot Grasp Detection}\label{subsec_zeroshot_exp}

\textbf{Setup.} We train three deep-learning grasp networks: GR-ConvNet~\cite{kumra2020antipodal}, Det-Seg-Refine~\cite{ainetter2021end}, and GG-CNN~\cite{morrison2018closing} on five datasets: Grasp-Anything, Jacquard~\cite{depierre2018jacquard}, Cornell~\cite{jiang2011efficient}, VMRD~\cite{zhang2019roi}, and OCID-grasp~\cite{ainetter2021end}. The primary metric is the success rate, defined similarly to~\cite{kumra2020antipodal}, requiring an IoU score above $0.25$ with the ground truth grasp and an offset angle less than $30^\circ$. For zero-shot learning's base and new labels~\cite{zhou2022conditional}, we utilize the classified LVIS's labels from Section~\ref{subsec: data-stat}. Initially, we merge LVIS labels from all five datasets and then identify the top $70\%$ labels by occurrence. These labels form the `Base' classes, while the remaining $30\%$ become `New' classes. We use the harmonic mean (`H') to measure the overall zero-shot success rates as in~\cite{zhou2022conditional}.


\textbf{Base-to-New Generalization.} We report the base-to-new grasp detection results in Table~\ref{table: base-to-new-grasp-detection}. There are two central observations from the results. First, three baselines GR-ConvNet, Det-Seg-Refine, and GG-CNN exhibit satisfactory performances over five datasets, implying there is less room for model-centric approaches to improve the grasp detection results on each separated dataset. Second, Grasp-Anything is more challenging to train as our detection results are lower than related datasets using the same approaches due to the greater coverage of unseen objects in the testing phase.

\begin{table}[htp]
\vspace{-1.5ex}
\centering
\renewcommand
\tabcolsep{4.5pt}
\hspace{1ex}
\caption{\label{table: cross-dataset} Cross-dataset result using GR-ConvNet~\cite{kumra2020antipodal} as the baseline.}
\vskip 0.1 in
\resizebox{\linewidth}{!}{
\begin{tabular}{@{}rccccc@{}}
\toprule
 \diagbox{Train}{Test} & \textbf{Jacquard} & \textbf{Cornell} & \textbf{VMRD} & \textbf{OCID-grasp} & \textbf{Grasp-Anything}\\
\midrule Jacquard~\cite{depierre2018jacquard} & \underline{0.89} & 0.51 & 0.13 & 0.21 & \textbf{0.17} \\

Cornell~\cite{jiang2011efficient} & 0.06 & \underline{0.98} & 0.20 & 0.12 & 0.15 \\

VMRD~\cite{zhang2019roi} & 0.06 & 0.21 & \underline{0.79} & 0.11 & 0.12 \\

OCID-grasp~\cite{ainetter2021end} & 0.09 & 0.12 & 0.20 & \underline{0.74} & 0.12 \\ \midrule

Grasp-Anything (ours) & \textbf{0.38} & \textbf{0.60} & \textbf{0.32} & \textbf{0.37} & \underline{0.67} \\
\bottomrule
\end{tabular}}
\vspace{-0.5ex}
\end{table}

\textbf{Cross-dataset Transfer Learning.} Table~\ref{table: cross-dataset} presents the results of training GR-ConvNet on a dataset (row) and testing on another dataset (column). For instance, when a GR-ConvNet is trained on Jacquard and tested on Cornell, an accuracy of 0.51 is achieved. Our dataset improves in about $9-29\%$ over other datasets. Particularly, when testing on Jacquard, the performance of GR-ConvNet trained on Grasp-Anything is four times better than other datasets.

\begin{table}[!ht]
\vspace{-1.5ex}
\centering
\renewcommand
\tabcolsep{4.5pt}
\hspace{1ex}
\caption{\label{table: generalized-zero-shot} Generalized Zero-shot Grasp Detection}
\vskip 0.1 in
\resizebox{\linewidth}{!}{\begin{tabular}{@{}rcccccccccccc@{}}
\toprule
& & \multicolumn{3}{c}{\textbf{Cornell}} & \multicolumn{3}{c}{\textbf{VMRD}} & \multicolumn{3}{c}{\textbf{OCID-Grasp}}\\
\cmidrule(lr){3-5}\cmidrule(lr){6-8}\cmidrule(lr){9-11}
Baseline & Dataset Used & Base &  New & H & Base &  New & H & Base &  New & H\cr 
\midrule
GR-ConvNet~\cite{kumra2020antipodal} & Jacquard & 0.53 & 0.50 & 0.51 & 0.18 & 0.10 &  0.13 & 0.23 & 0.19 & 0.21 \\ 
GR-ConvNet~\cite{kumra2020antipodal} & Grasp-Anything & \textbf{0.68} & \textbf{0.57} & \textbf{0.62} & \textbf{0.33} & \textbf{0.29} &  \textbf{0.31} & \textbf{0.42} & \textbf{0.37} & \textbf{0.39} \\ \midrule

Det-Seg-Refine~\cite{ainetter2021end} & Jacquard & 0.55 & 0.53 & 0.54 & 0.15 &  0.06 & 0.09 & 0.23 & 0.16  & 0.19 \\
Det-Seg-Refine~\cite{ainetter2021end} & Grasp-Anything & \textbf{0.72} & \textbf{0.61} & \textbf{0.66} & \textbf{0.30} &  \textbf{0.26} &
\textbf{0.28} &  \textbf{0.39} &  \textbf{0.31} &
\textbf{0.35} \\ \midrule

GG-CNN~\cite{morrison2018closing} & Jacquard & 0.45 & 0.39 & 0.42 & 0.10 &  0.07 & 0.08 & 0.18 & 0.10 & 0.12 \\
GG-CNN~\cite{morrison2018closing} & Grasp-Anything & \textbf{0.64} & \textbf{0.56} & \textbf{0.60} & \textbf{0.22} &  \textbf{0.18} &
\textbf{0.19} &  \textbf{0.36} &  \textbf{0.27} &
\textbf{0.31} \\

\bottomrule
\end{tabular}}
\vspace{-0.5ex}
\end{table}

\textbf{Generalized Zero-shot Learning.} 
We conduct a generalized zero-shot grasp detection in Table~\ref{table: generalized-zero-shot}. We compare our dataset with Jacquard, as it shows competitive results in Tables~\ref{table: base-to-new-grasp-detection} and~\ref{table: cross-dataset}. The other three datasets are used for testing. In all scenarios, Grasp-Anything considerably improves the performances of all baselines.


\subsection{Robotic Evaluation}\label{sub_sec_robot_exp}

\begin{figure}[t]
\centering
\subfigure{
\label{fig:rotbot1}
\def\svgwidth{1\columnwidth}
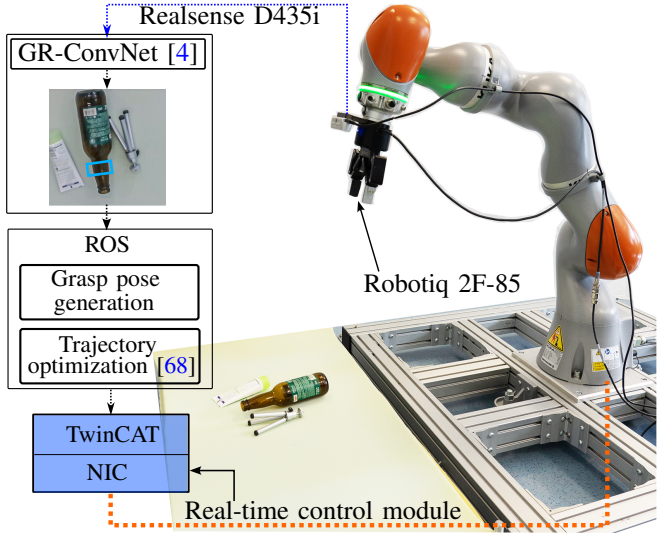
}
\vspace{1pt}
\caption{\textbf{Overview of the robotic experiment setup.}}
\label{fig: robot demonstration}
\vspace{-0.2ex}
\end{figure}

\textbf{Setup.} Our robotic evaluation on a KUKA robot is shown in Figure~\ref{fig: robot demonstration}. We use GR-ConvNet~\cite{kumra2020antipodal} as the grasp detection network. The grasp detection results are converted into a 6DOF grasp pose using the depth image from a RealSense camera, as in~\cite{kumra2020antipodal}. The trajectory planner \cite{vu2023machine,beck2022singularity} is used to execute the grasp. The evaluation is conducted for a single object and a cluttered scenario utilizing a set of $15$ objects. We conduct experiments for each scenario $60$ times with network weights of GR-ConvNet~\cite{kumra2020antipodal} trained on the proposed Grasp-Anything, OCID-grasp, VMRD, and Jacquard datasets, respectively. 

\textbf{Results.} Table \ref{table: real-world-robot} statically shows the effectiveness of our proposed dataset, which helps the GR-ConvNet achieve the highest performance ($93.3\%$ for the single object and $91.6\%$ for the cluttered scene) compared to other datasets. Several demonstrations can be found in our Demonstration Video.

\begin{table}[t]
\centering
\renewcommand
\tabcolsep{4.5pt}
\hspace{1ex}
\caption{\label{table: real-world-robot} Real-world Grasping Experiments.}
\vskip 0.1 in
\begin{tabular}{@{}crcc@{}}
\toprule
Baseline & Dataset Used & Single Object &  Cluttered Scene\cr 
\midrule
\multirow{5}{*}{\rotatebox[origin=c]{90}{\scalebox{0.8}{GR-ConvNet~\cite{kumra2020antipodal}}}} & 
Cornell & 81.6\% & 68.3\%\\ 
& Jacquard & 85.0\% & 88.3\% \\ 
& VMRD & {78.3\%} & {70.0\%}\\
& OCID-grasp & {80.0\%} & {71.6\%}\\
& Grasp-Anything  (ours) & \textbf{93.3\%} & \textbf{91.6\%}\\
\bottomrule
\end{tabular}
\end{table}


\subsection{Grasp Detection in the Wild}
\begin{figure}[!ht]
    \centering
    \includegraphics[width=\linewidth]{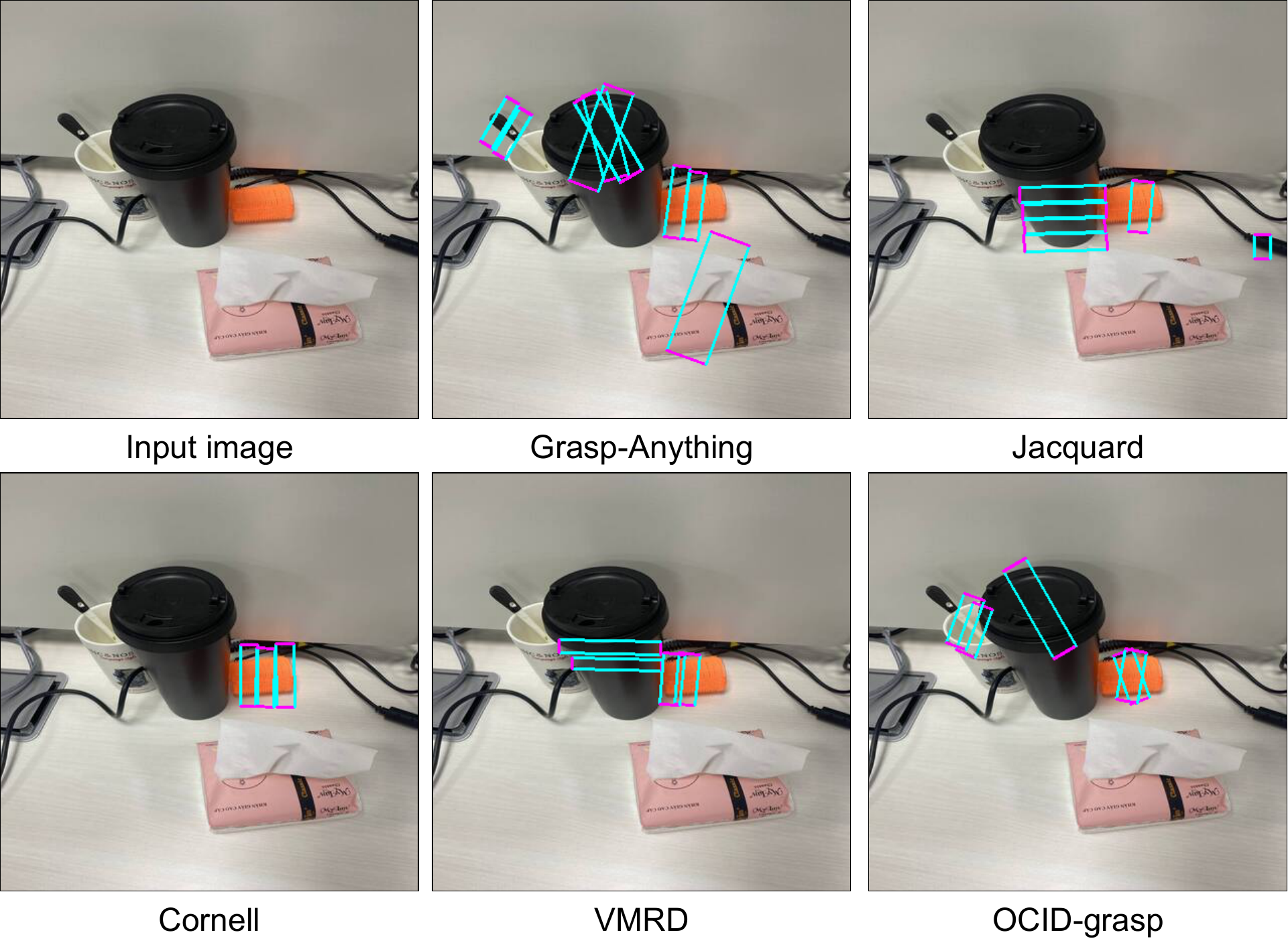}
    \caption{\textbf{Qualitative results.} We use GR-ConvNet across different datasets.}
    \label{fig:qualitative-result}
\end{figure}

\begin{figure}[!ht]
    \centering
    \includegraphics[width=\linewidth]{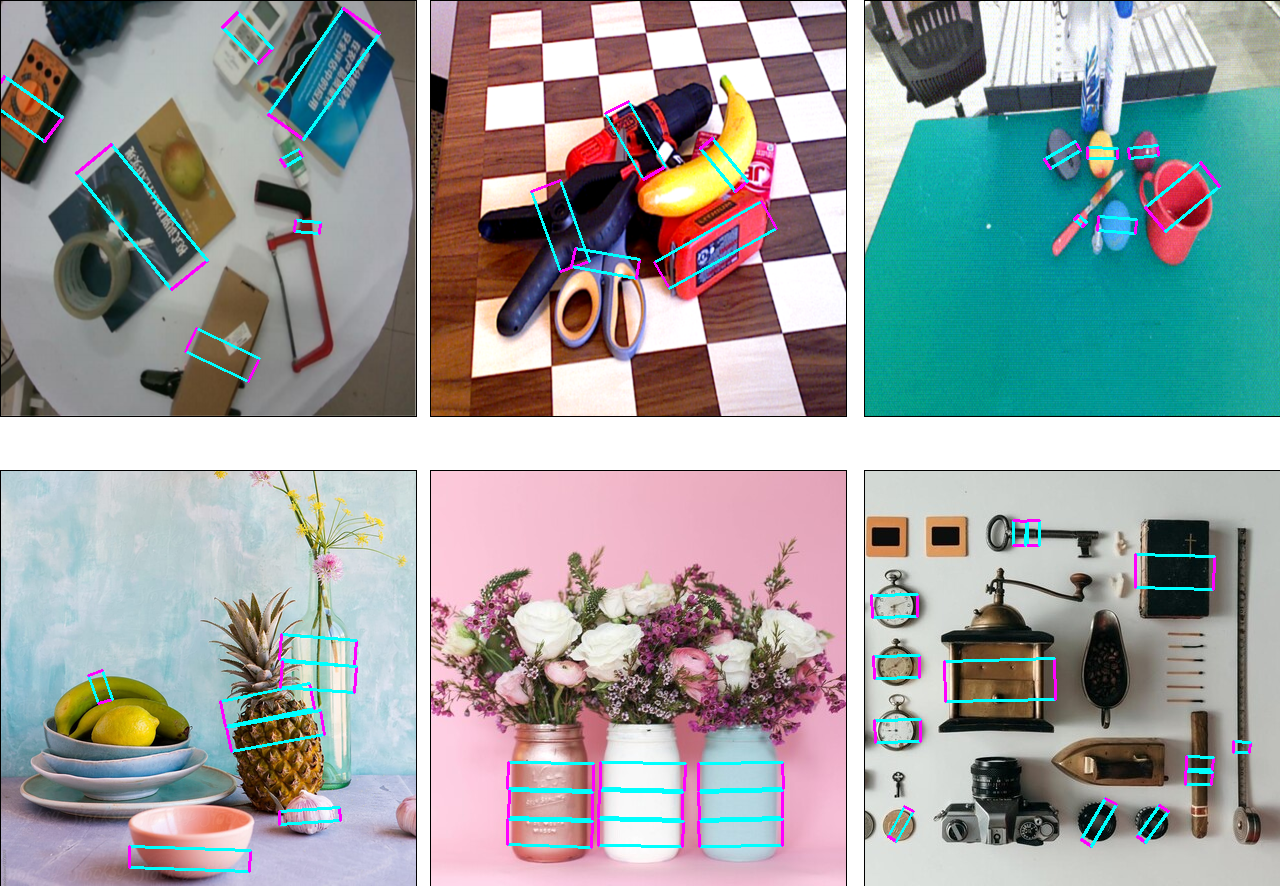}
    \vspace{-1ex}
    \caption{\textbf{In the wild grasp detection with a model trained on our dataset.} The top row are images from other datasets: NBMOD~\cite{cao2023nbmod}, YCB-Video~\cite{xiang2017posecnn}, GraspNet~\cite{fang2020graspnet}; while the bottom row includes internet images. 
    }
    \label{fig:in-the-wild}
    \vspace{-1ex}
\end{figure}

We visualize grasp detection of a daily office arrangement image from GR-ConvNet trained by different datasets in Fig.~\ref{fig:qualitative-result}. This figure shows that  Grasp-Anything can improve the grasp detection quality over related datasets when the same baseline network is used. Fig.~\ref{fig:in-the-wild} showcases grasp detection examples utilizing a pretrained GR-ConvNet on the Grasp-Anything dataset on random images from the internet and different datasets. We can see that the detected grasp poses are adequate in quality and quantity. 

\subsection{Discussion}
Through the zero-shot experiment (Section~\ref{subsec_zeroshot_exp}), we can answer our first question and conclude that our Grasp-Anything can be used as a challenging dataset for grasp detection as networks trained on our dataset achieve lower accuracy compared to other datasets (Table~\ref{table: base-to-new-grasp-detection}). Furthermore, the cross-dataset experiment in Table~\ref{table: cross-dataset} implies that utilizing our dataset on other datasets brings significant improvement. Also, the robotic experiments (Section~\ref{sub_sec_robot_exp}) demonstrate that the model trained on our synthesis dataset outperforms the same model trained on different datasets with real-world images. This answers our second question and validates the usefulness of our dataset in real-world robotic experiments.

While promising results have been achieved with our dataset, we see several important improvement points. First, we remark that our dataset's creation is time-consuming and relies on access to the commercial ChatGPT API. Specifically, it took approximately three months to generate and process 1M scene descriptions on a cluster of three NVIDIA Quadro 8000. Fortunately, future research can reuse our provided assets (images, prompts, etc.) without starting from scratch. Second, our dataset currently lacks 3D point clouds. This is primarily because text-to-point-cloud or image-to-point-cloud foundation models have yet to achieve convincing results~\cite{deitke2023objaverse-xl}. Therefore, creating point clouds from our prompts and images would make our dataset more useful in robotic tasks. Finally, since our dataset includes text prompts, we believe it will foster interesting research directions, such as language-driven grasping and human-robot interaction.

\section{Conclusions}\label{Sec: conclusion}
We have presented Grasp-Anything, a new large-scale language-driven dataset for robotic grasp detection. Our analyses demonstrate that Grasp-Anything encompasses many objects and natural scene arrangements. 
The experiments on different networks and datasets with the real robot reveal that our Grasp-Anything dataset improves significantly over related datasets. 
By incorporating natural scene descriptions, we hope our dataset can serve as a foundation dataset for language-driven grasp detection.

\bibliographystyle{IEEEtran}
\bibliography{IEEEabrv, reference}
   
\end{document}